\documentclass[sigconf,authorversion,nonacm]{acmart}

\usepackage{multirow}
\usepackage{float}
\AtBeginDocument{%
  }

\begin{document}

\title{Multiscale Contextual Learning for Speech Emotion Recognition in Emergency Call Center Conversations}
\author{Théo Deschamps-Berger}
\email{theo.deschamps@lisn.fr}
\orcid{1234-5678-9012}
\affiliation{%
  \institution{LISN-CNRS, Paris-Saclay University}
  \country{France}
}

\author{Lori Lamel}
\email{lamel@lisn.fr}
\affiliation{%
  \institution{LISN-CNRS, Paris-Saclay University}
  \country{France}
}

\author{Laurence Devillers}
  \email{devil@lisn.fr}
\affiliation{%
  \institution{LISN-CNRS, Sorbonne-University}
  \country{France}
}


\begin{abstract}

Emotion recognition in conversations is essential for  
ensuring advanced human-machine interactions. However, creating robust and accurate emotion recognition systems in real life is challenging, mainly due to the scarcity of emotion datasets collected in the wild and the inability to take into account the dialogue context. The CEMO dataset, composed of conversations between agents and patients during emergency calls to a French call center, fills this gap. The nature of these interactions highlights the role of the emotional flow of the conversation in predicting patient emotions, as context can often make a difference in understanding actual feelings. This paper presents a multi-scale conversational context learning approach for speech emotion recognition, which takes advantage of this hypothesis. We investigated this approach on both speech transcriptions and acoustic segments. Experimentally, our method uses the previous or next information of the targeted segment. 
In the text domain, we tested the context window using a wide range of tokens (from 10 to 100) and at the speech turns level, considering inputs from both the same and opposing speakers. According to our tests, the context derived from previous tokens has a more significant influence on accurate prediction than the following tokens. Furthermore, taking the last speech turn of the same speaker in the conversation seems useful. In the acoustic domain, we conducted an in-depth analysis of the impact of the surrounding emotions on the prediction. While multi-scale conversational context learning using Transformers can enhance performance in the textual modality for emergency call recordings, incorporating acoustic context is more challenging.

\end{abstract}

\ccsdesc[500]{Computing methodologies~Discourse, dialogue and pragmatics}
\ccsdesc[500]{Computing methodologies~Natural language processing}

\keywords{Speech emotion recognition, Multiscale contextual learning, Emotion Recognition in Conversation, Transformers, Emergency call center}
\begin{teaserfigure}
  \includegraphics[width=\textwidth]{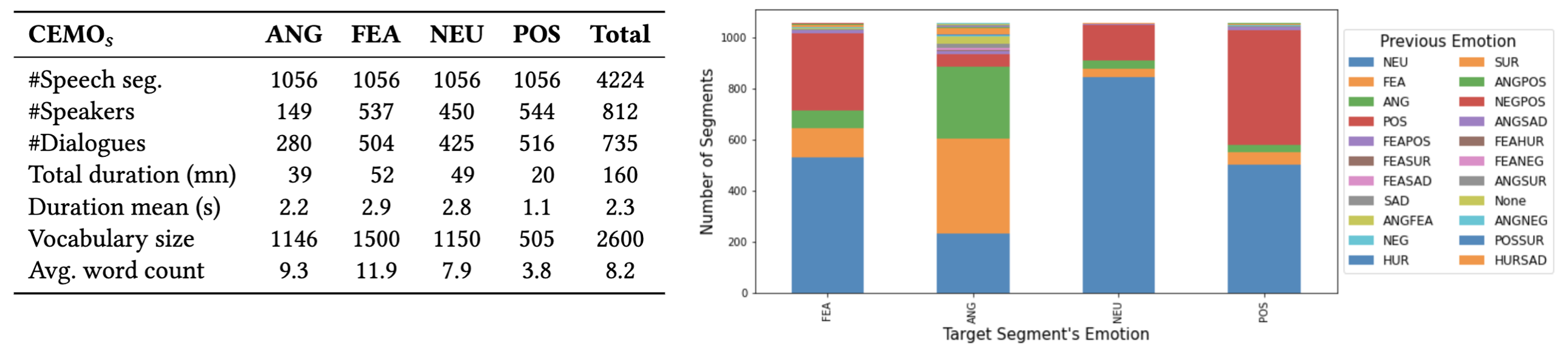}
  \caption{The Multiscale Contextual Learning has been applied for Speech Emotion Recognition on a French emergency call center conversations corpus named CEMO \cite{devillersChallengesReallifeEmotion2005a}. A subset of this corpus with 4 emotions (Anger, Fear, Neutral and Positive state), described in the left table has been used for this contextual study. The right figure shows the distribution of the previous emotion segments for the 4 target emotions.}
  \Description{}
  \label{fig:teaser}
\end{teaserfigure}

\maketitle

\section{Introduction and recent work}

In recent years novel methods and techniques have been applied to speech-based downstream applications with a focus on the potential benefits of incorporating conversational information into such systems. This contextual information is usually derived from previous and subsequent utterances in the form of speech transcriptions or acoustic contexts. 

An early significant approach by \cite{poriaContextDependentSentimentAnalysis2017}, utilized a bidirectional LSTM to assimilate context without distinguish speakers. Extending this methodology, \cite{hazarikaICONInteractiveConversational2018} incorporated a GRU structure within their ICON model to identify speaker's relationships. 
Later, \cite{ghosalDialogueGCNGraphConvolutional2019}, converted conversations into a graph, employing a graph convolutional neural network for emotion classification. This work was further developed by (almost) the same team, who integrated common-sense knowledge to understand interlocutors' interactions \cite{ghosalCOSMICCOmmonSenseKnowledge2020}.

Recent work by \cite{shenDirectedAcyclicGraph2021} has used new neural network structures for context understanding. 
An extension of this approach was proposed in \cite{huDialogueCRNContextualReasoning2021} which introduced DialogueCRN to fully capture conversational context from a cognitive point of view. 
These papers illustrate the ongoing evolution of the field.

Ongoing research about conversational context in speech task has paralleled the rise of self-supervised pre-training models, which are now popular for handling downstream tasks. These models has shown strong results across various speech tasks benchmarks as highlighted in \cite{wagnerDawnTransformerEra2022}. 
Our paper proposes context-aware fine-tuning, which utilizes surrounding speech segments during fine-tuning to improve performance on downstream speech tasks and enrich Transformer embeddings through the integration of auxiliary context module, as illustrated by \cite{shonContextAwareFineTuningSelfSupervised2023} and by \cite{liEmoCapsEmotionCapsule2022} with their emotion-aware Transformer Emoformer.

In the field of Speech Emotion Recognition, advances with Transformer models in deep learning have reached state-of-the-art performance on acted speech \cite{linSurveyTransformers2022} and on widely-known 
open-source research database like \cite{bussoIEMOCAPInteractiveEmotional2008}. Upon appropriate fine-tuning Transformers are able to learn efficient representations of the inputs.

However recognizing spontaneous emotions remains a challenge. But remarkably, Transformer encoder models shown significant results over classical approaches on spontaneous emotion recordings \cite{deschamps-bergerEndtoEndSpeechEmotion2021}. Through a specific integration of multimodal fusion mechanisms, these models are highly capable of gathering efficient emotional cues across modalities, \cite{deschamps-bergerExploringAttentionMechanisms2023}. This paper leverages the French CEMO corpus which consists of real-life conversational data collected in an emergency call center \cite{devillersChallengesReallifeEmotion2005a}. This corpus provides an excellent opportunity to tackle the challenge of integrating conversation context in a realistic emergency context.

Despite the effectiveness of Transformer models, their standard self-attention mechanism's quadratic complexity limits application to relatively small windows [3]. Cutting-edge research has focused on optimizing the attention mechanisms to a lower complexity like FlashAttention \cite{daoFlashAttentionFastMemoryEfficient2022}, addressing this limitation by lowering the attention complexity paves the way for future models to be trained from scratch on huge datasets with wider context.

In this work we propose a multi-scale hierarchical training system adapted to pre-trained standard attention models which are available by the French community. The proposed approach draws inspiration from recent work by \cite{shonContextAwareFineTuningSelfSupervised2023}.
We evaluate the impact of different types of contextual information for acoustic level and manual speech transcription. Integrating the acoustic and linguistic context of dialogue into an emotion detection system remains a challenge, but this work aims to contribute to these ongoing efforts and explain the impact of such a system and their limitations.

\section{Conversational corpus: CEMO }

The emergency call center corpus presents a unique opportunity to examine real-world emotional expression. This rich 20+ hour dataset captures naturalistic interactions between callers in crisis and operators. As described by \cite{devillersChallengesReallifeEmotion2005a, vidrascuDetectionReallifeEmotions2005}, it contains emotional annotations across situations ranging from medical emergencies to psychiatric distress. Segments were coded for major and minor emotions with fine-grained labels from 7 macro-classes.

\begin{table}
\centering
\caption{The 10 most represented emotions and mixtures of emotions by caller and agent.{ FEA: Fear, NEU: Neutral, POS: Positive, ANG: Anger, SAD: Sadness, HUR: Hurt, SUR: Surprise OTHER}: Sum of remaining classes}
\label{tab:1}
\resizebox{\columnwidth}{!}{%
\begin{tabular}{lrrllrr} 
\toprule
\textbf{Caller} & \textbf{Segments} & \textbf{Speakers} &  & \textbf{Agent} & \textbf{Segments} & \textbf{Speakers}  \\ 
\midrule
Total            & 17679             & 870\hphantom{0}               &  & Total          & 16523             & 7\hphantom{00}                  \\
FEA              & 7397              & 825\hphantom{0}               &  & NEU            & 10059             & 7\hphantom{00}                   \\
NEU              & 7329              & 822\hphantom{0}               &  & POS            & 4310              & 7\hphantom{00}                   \\
POS              & 1187              & 566\hphantom{0}               &  & ANG            & 1213              & 6\hphantom{00}                   \\
ANG              & 417               & 146\hphantom{0}               &  & FEA            & 437               & 7\hphantom{00}                   \\
HUR              & 261               & 67\hphantom{0}                &  & FEA/POS        & 122               & 4\hphantom{00}                   \\
SUR              & 144               & 118\hphantom{0}              &  & ANG/POS        & 65                & 4\hphantom{00}                   \\
FEA/POS          & 130               & 103\hphantom{0}              &  & ANG/FEA        & 57                & 3\hphantom{00}                   \\
FEA/SAD          & 128               & 71\hphantom{0}                &  & POS/SUR        & 24                & 4\hphantom{00}                   \\
FEA/HUR          & 116               & 55\hphantom{0}                &  & FEA/SUR        & 16                & 4\hphantom{00}                   \\
OTHER            & 294               & 171\hphantom{0}              &  & OTHER          & 52                & 3\hphantom{00}                   \\
\bottomrule
\end{tabular}
}
\end{table}

The caller can be either the patient or a third party (family, friend, colleague, neighbor, stranger). The wide range of caller types (age, gender, origin), accents (regional, foreign), different vocal qualities (alterations due to alcohol/medication, a cold, etc.) also makes it an extremely diverse corpus. As shown in Table 1, the Caller and Agent emotional profiles differ. Callers expressed intense emotions like fear, anger, and sadness, given their crisis state. In contrast, agents maintained a regulated presence, with more positive and neutral states, reflecting their professional role.

Inter-rater reliability highlights differences between callers and agents. Agreement on emotions was higher for callers than agents (Kappa 0.54 vs 0.35). This suggests agents regulate emotions, producing subtle expressions that are challenging to consistently code. Refining annotation schemes could better capture the complexity of agents' emotional states.

\begin{table}
\centering
\caption{Details of the CEMO subset of speech signals and manual transcripts. ANG: Anger, FEA: Fear, NEU: Neutral, POS: Positive, Total: Total number of segments.}
\label{tab:CEMOdetails}
\resizebox{\columnwidth}{!}{
\begin{tabular}{lccccc} 
\toprule
\textbf{CEMO$_s$}           & \textbf{ANG}  & \textbf{FEA}  & \textbf{NEU}  & \textbf{POS}  & \textbf{Total}  \\ 
\midrule
\#Speech seg.    & 1056 & 1056 & 1056 & 1056 & 4224   \\
\#Speakers           & 149  & 537  & 450  & 544  & 812    \\
\#Dialogues          & 280  & 504  & 425  & 516  & 735    \\
Total duration (mn)    & \hphantom{0}39   & \hphantom{0}52   & \hphantom{0}49   & \hphantom{0}20   & 160    \\
Duration mean (s) & 2.2  & 2.9  & 2.8  & 1.1  & 2.3    \\
Vocabulary size    & 1146 & 1500 & 1150 & 505  & 2600   \\
Avg.\ word count & 9.3  & 11.9 & 7.9  & 3.8  & 8.2    \\
\bottomrule
\end{tabular}
}
\end{table}

Data preparation is key for performance and robustness. As detailed in Table \ref{tab:CEMOdetails}, a balanced CEMO subset (2h40) of 4224 segments was selected for training/validation/testing. The 4 classes were equally distributed with 1056 samples each. Fear and Neutral were subsampled, prioritizing speaker diversity. Anger was completed with agent segments of annoyance/impatience resulting in a class with less speakers diversity and possible bias. Positive had the most speakers and dialogues, suggesting heterogeneity. Manual transcriptions were performed with guidelines similar to the Amities project \cite{hardyMultilayerDialogueAnnotation2003}. 
 
The transcriptions contain about 2499 nonspeech markers, primarily pauses, breath, and other mouth noises. The vocabulary size is 2.6k, with a mean and median of about 10 words per segment (min 1, max 47).

\begin{figure}[h]
    \centering
    \includegraphics[width=0.4\textwidth]{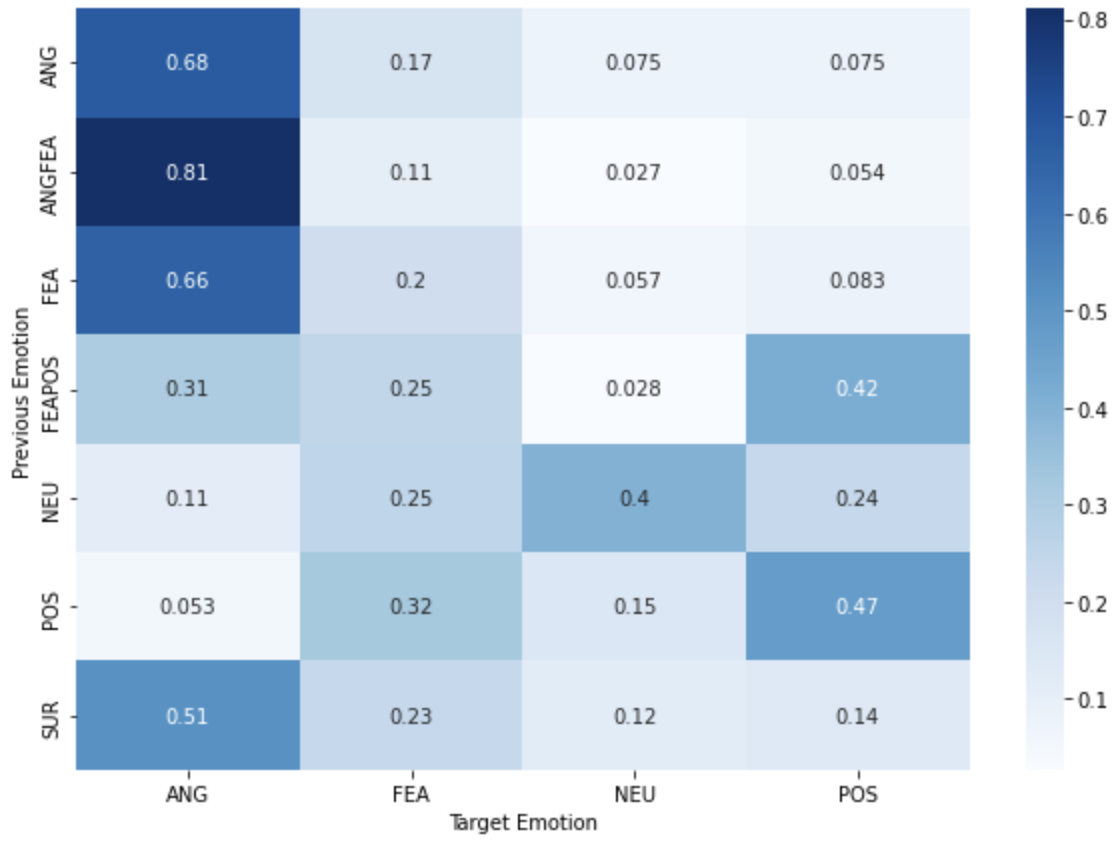}
    \caption{Transition between the previous and current emotion segments, we only show previous emotions which have at least 30 segments.}
    \label{fig:trans_between_emos}
\end{figure}

Figure \ref{fig:trans_between_emos} represents the transition probabilities between the emotion expressed in the previous speech turn and the target segment. The diagram illustrates the likelihood of moving from each prior emotion category (rows) to each target emotion (columns). Anger persists across turns at a 68\% probability. Asymmetry exists between Anger and Fear, with Fear more often following Anger. Surprise is surprisingly followed by Anger, without any wordplay intended.


\begin{figure}[h]
    \centering
    \includegraphics[width=0.47\textwidth]{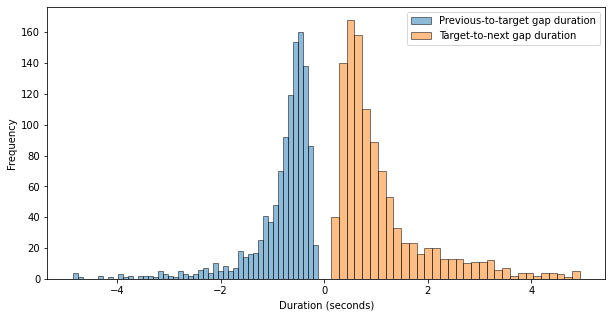}
    \caption{Histogram of gap duration between context segment and the segment to predict (segments with a gap of zero are excluded)
    } %
    \label{fig:time_gap}
\end{figure}

Figure \ref{fig:time_gap} displays a histogram which illustrates the distribution of gap duration between the context and the target segment. This excludes contiguous segments, corresponding to 3040 Previous-to-target and 2895 Target-to-next segments. For non-contiguous segments, there are 1174 and 1152 respectively for Previous-to-target and Target-to-next segments. Notably, there are only 10 segments that lack any preceding context, and 177 segments that do not have any following context.

\section{Methodology}

\par Our approach aims at recognizing emotions from speech. The systems presented in this article are based on the incorporation of conversational context via pre-trained transformative attention mechanisms. We have divided this section into two main parts, devoted to single modalities (acoustic and textual). Our aim is to better understand the impact of context in these systems.

\par First, we tackled the textual modality, i.e. manual transcriptions of dialogues incorporating the context in a "blind" way a defined number of conversational elements (named tokens in pre-trained models). Then, we modified the scale of the contextual window as a function of speech turns, and conducted experiments on specific conversational segments.

\par In a second phase, we focused on the acoustic modality, where we exploited the context of speech turns that had been supported by the textual approach. We then extended this to hierarchical training, on the assumption that low-level cues for emotion prediction would be learned by the model during initial context-free training, and that incorporating conversational context in a second phase would enable higher-level information to be learned.

\par Our methodology is based on the application of specific Transformer encoder models: FlauBERT large \cite{leFlauBERTUnsupervisedLanguage2020a} and wav2vec2.0 large \cite{baevskiWav2vecFrameworkSelfSupervised2020b}. These models use self-supervised learning to create meaningful abstractions from text and raw audio data. Prior research \cite{deschamps-bergerInvestigatingTransformerEncoders2022} showed the successful adaptation of pre-trained models to detect discrete speech emotion labels from the CEMO corpus \cite{devillersChallengesReallifeEmotion2005a}. From the available models, we chose to use the leBenchmark model (Wav2Vec2-FR-3K)~\cite{evainLeBenchmarkReproducibleFramework2021a}, trained on 3,000 hours of French language data. This decision was guided by the model's performance on the CEMO corpus \cite{deschamps-bergerInvestigatingTransformerEncoders2022}.

The training database for the wav2vec2-FR-3K model is comprised of spontaneous dialogues recorded by telephone, some with emotional content, thus mirroring the characteristics of the CEMO corpus. The multi-head attention layers were fine-tuned for speech emotion recognition using the CEMO corpus. This was done under the assumption that the initial layers of the model (Convolutional layers and Embedding) are robust to this task \cite{wagnerDawnTransformerEra2022, deschamps-bergerInvestigatingTransformerEncoders2022}.

\section{Contextual Exploration of Textual Modality}

In this research, we propose a fine-tuned system for detecting emotions on the CEMO dataset by incorporating semantic information from the anterior or posterior parts of speech. During training, the context is concatenated with speech inputs to be fed into a Transformer. The proposed system relies on the pre-trained multi-head attention layers of the FlauBERT model \cite{leFlauBERTUnsupervisedLanguage2020a}, to learn the relationships between the latent states of the current segment and its context. 

The multi-head attention mechanism allows the model to learn relevant parts of the segment to predict, within its conversational context. To emphasis this weighting, we mask the embeddings yielded by the Transformer corresponding to the context. The rest of the embeddings are fed into an attention pooling layer and classified into discrete emotions.



\begin{figure}[h]
    \centering
    \includegraphics[width=0.40\textwidth]{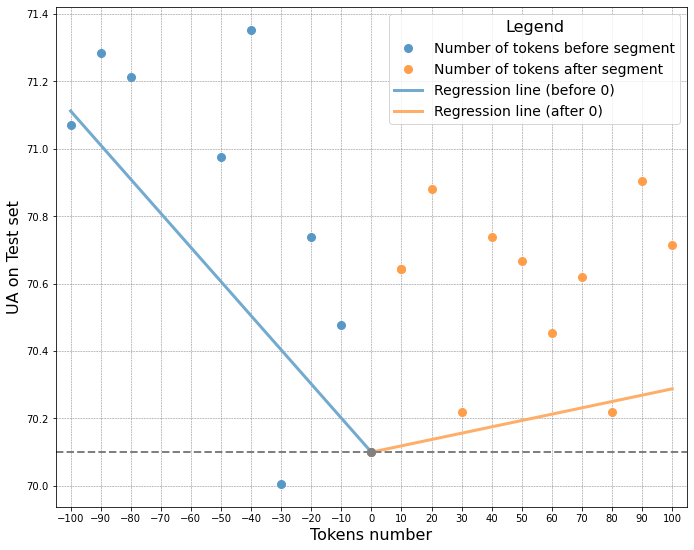}
    \caption{Prediction Accuracy vs. Context Token Count: Tokens number represent anterior/posterior tokens to the target segment. Accuracy is Unweighted Average (UA), in \%}
    \label{fig:enter-label}
\end{figure}

We firstly focused on a "blind" semantic approach where the context was selected by the amount of tokens. The average number of seconds for one token in the CEMO dataset is equivalent to 0.2s, then we have an average of 5 tokens per second. We performed some experiments with a window of tokens' number from 0 to 100. The results are displayed in the Figure. \ref{fig:context} which shows the UA scores obtained in the prediction of the four discrete emotions. Two regression lines pass through the origin 0 which correspond to the baseline experiment without context. 

There is a positive impact of context unevenly distributed between the anterior and posterior conversational contexts. The previous tokens in our tokens are more useful to enrich the segment embeddings to be predicted.
Limits to the interpretability of this approach may arise from the semantic perspective, where we are uncertain whether the number of tokens will be extracted from the middle of a sentence or a speech turn. 

To address this hypothesis, we conducted experiments at the speech turn level, using the previous or next segment of speech. We also extended the experiments to speaker type, which could have an impact on how the context is learned by the Transformer.

\begin{table}
\centering
\caption{
{Comparison of Textual Models (on manual transcriptions) using FlauBERT Embeddings with and without Contextual Information, sorted by UA: \%} Contextual information: 1st column: Previous or Next segments, 2nd column: same speaker, opposite speaker or all speakers}
\label{tab:context}
\resizebox{\linewidth}{!}{%
\begin{tabular}{clllllll} 
\toprule
\multicolumn{1}{l}{\textbf{Model}} & \textbf{Context} & \textbf{from} & \textbf{ANG} & \textbf{FEA} & \textbf{NEU} & \textbf{POS} & \multicolumn{1}{c}{\textbf{Total}}  \\ 
\midrule
\multirow{6}{*}{FlauBERT}
& Previous & same speaker      & 66.0  & 64.5 & 70.6 & 85.7  & \textbf{71.7}   \\
& Next & same speaker          & 70.4  & 59.7 & 72.5 & 83.6  & \textbf{71.5}   \\
& Next & opposite speaker      & 67.9  & 62.3 & 72.4 & 82.7  & \textbf{71.3}   \\
& Previous &      all speakers & 66.3  & 61.2 & 72.4 & 84.8  & \textbf{71.2}   \\
& Next & all speakers          & 64.1  & 66.3 & 68.6 & 85.2  & \textbf{71.0}   \\
& Previous & opposite speaker  & 59.4  & 66.1 & 71.0 & 84.3  & \textbf{70.2}   \\
\midrule
FlauBERT &    Without       &- & 61.1  & 66.0 & 68.2 & 85.1  & \textbf{70.1}   \\
\bottomrule
\end{tabular}
}
\end{table}

The results in Table. \ref{tab:context} detailed the different configurations we used. From the results, it seems that incorporating context from the same speaker outperforms the opposite speaker approach, suggesting that the emotion of a sentence may be more influenced by the speaker's previous sentences rather than the other speaker's. This makes intuitive sense as people's emotions tend to be consistent within a short time frame and are likely to be less influenced by the immediate response from others. The Anger and Fear classes fluctuate the most with context, which may indicate that these emotional states are more complex or nuanced, and may be more influenced by context and speaker.

\par Contextual experiments on the speech turns scale produced better or similar results to those obtained on the token scale, see Fig. \ref{fig:context} and Table. \ref{tab:context}. Even with a large token window, up to 100 tokens (sub-words for FlauBERT), equivalent to around 20 seconds of speech, it fails to achieve the best scores, regardless the turn before or after the segment to be predicted. 

\par In comparison the average context speech turn segments last 1 seconds, thus, the right positioning and semantic meaning of the text is one of Speech Emotion Recognition's keys performance.

\section{Contextual Exploration of Acoustic Modality}

\par Our approach to predict emotions from acoustic is similar to the text modality, we concatenate raw audio as input to the acoustic Transformer and mask the embeddings specific to the context produced by the Transformer. At this stage, the wav2vec2 model applies a multi-head attention mechanism on both the surrounding segments and the target segment.

\begin{figure}[h]
    \centering
    \includegraphics[width=0.4\textwidth]{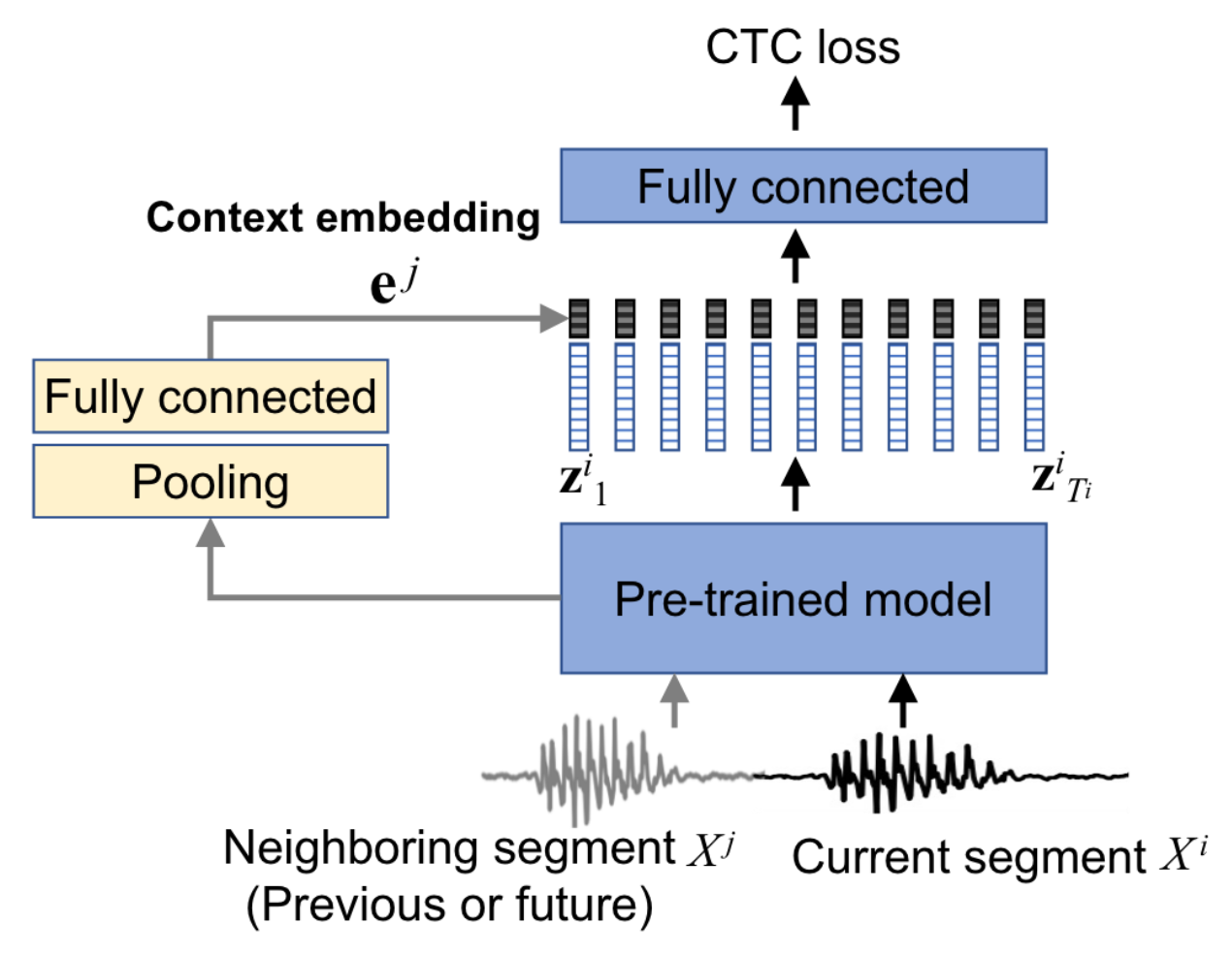}
    \caption{Illustration of CCFTE Concatenation of Context Features with Target Embeddings, figure from \cite{shonContextAwareFineTuningSelfSupervised2023}}
    \label{fig:context}
\end{figure}
 
\par This mechanism allows the model to focus on different features in the segment and its surrounding context, potentially improving the emotional relevance of the embeddings produced. 

\par To adjust the wav2vec2-FR-3K model to our needs, we added an attention pooling layer and a classifier.  
One drawback of this approach is the higher computational cost of the Transformer acoustic model compared to the textual one. Due to the specifications of our computational clusters, we are limited to a maximum length of around 6.5 seconds for the large wav2vec2 model.

\par Following the results obtained on context at a speech turns level with the text modality, we incorporate the context from the previous or next turn of the target segment. Furthermore we implemented a novel way to enrich the yielded wav2vec2 embeddings through a dedicated auxiliary context module influenced by \cite{shonContextAwareFineTuningSelfSupervised2023}. 
The auxiliary module is detailed in Fig. \ref{fig:context}, it gathers the embeddings from the surrounding segments into a context attention pooling layer. This pool, together with a fully connected network, generates a context vector that provides a compact, informative representation of the surrounding context.

\begin{equation}
C_{i} = \text{FullyConnected}( \text{AttentionPooling}(E_{i}, S_{i}))
\end{equation}

In the equation above, $C_{i}$ is the context vector for the $i$-th segment, $E_{i}$ signifies the embeddings of the target segment, and $S_{i}$ is the input segment. The context vector is then concatenated with each of the embeddings of the target segment, effectively underline the contextual information into the final classifier prediction.

\begin{table}
\caption{Comparison of Acoustic Models Using wav2vec2 Embeddings with and without Contextual Information, MWCE: Masking w2v2 context embed., CCFTE: Concatenation of Context Features with Target Embeddings, sorted by UA}
\resizebox{\columnwidth}{!}{%
\begin{tabular}{@{}lllllllllllll@{}}
\toprule
\textbf{Model}  & \textbf{Strategy}              & \textbf{Context}  & \textbf{MWCE} & \textbf{CCFTE} & \textbf{ANG}  & \textbf{FEA}  & \textbf{NEU}  & \textbf{POS}  & \textbf{Total} \\ \midrule
wav2vec2      & -             & Without  & -    & -   & 73.0 & 70.2 & 72.2 & 87.1 & \textbf{75.6} \\ \midrule
\multirow{4}{*}{wav2vec2}                & \multirow{4}{*}{Concatenation} & Previous & \checkmark    & \checkmark   & 71.1 & 73.1 & 70.7 & 86.6 & \textbf{75.4} \\
               &  & Next     & \checkmark    & \checkmark   & 68.1 & 73.7 & 73.1 & 86.4 & \textbf{75.3}  \\
               &  & Next     & \checkmark    & -   & 71.2 & 67.2 & 75.6 & 85.4 & \textbf{74.9} \\
               &  & Previous & \checkmark    & -   & 66.4 & 63.0 & 79.2 & 85.4 & \textbf{73.5} \\ \bottomrule
\end{tabular}%
}
\label{tab:audio_results}
\end{table}


The Table \ref{tab:audio_results} presents results evaluating the incorporation of contextual acoustic information to enhance emotion recognition performance of wav2vec2 embeddings. Across conditions, two proposed context integration methods were examined - masking the context embedding (MWCE) and concatenating context features with target embeddings (CCFTE) - using either previous or next utterances as context.

Notably, the baseline wav2vec2 model with no context elicited the highest total unweighted accuracy (UA) of 75.6\%, exceeding all context-enhanced models. This suggests intrinsic limitations of the concatenation-based context integration approaches assessed. Both MWCE and CCFTE concatenation utilizing prior context modestly boosted performance to 75.4\% UA. However, next context yielded negligible gains, indicating contextual benefits may be asymmetric.

\begin{table}
\caption{Hierarchical Training: Fine-tuning of Models with and without Context from a Baseline Checkpoint, MWCE: Masking w2v2 context embed., CCFTE: Concatenation of Context Features with Target Embeddings, sorted by UA}
\resizebox{\columnwidth}{!}{%
\begin{tabular}{@{}lllllllllllll@{}}
\toprule
\textbf{Model}         & \textbf{Strategy}      & \textbf{Context}  & \textbf{MWCE} & \textbf{CCFTE} & \textbf{ANG}  & \textbf{FEA}  & \textbf{NEU}  & \textbf{POS}  & \textbf{Total} \\ \midrule
wav2vec2      & -             & -        & -    & -     & 76.5 & 72.7 & 69.9 & 85.6 & \textbf{76.2}  \\ \midrule
 \multirow{4}{*}{wav2vec2}                & \multirow{4}{*}{Concatenation} & next     & \checkmark    & \checkmark     & 74.7 & 70.0 & 72.8 & 87.1 & \textbf{76.2}  \\
              &  & next     & \checkmark    & -     & 76.2 & 70.1 & 70.7 & 87.5 & \textbf{76.1}  \\
              &  & previous &  \checkmark    & -     & 73.6 & 71.5 & 72.1 & 86.6 & \textbf{75.9}  \\
              &  & previous &  \checkmark    &  \checkmark     & 75.5 & 70.4 & 70.9 & 85.4 & \textbf{75.5}  \\ \bottomrule
\end{tabular}
}
\label{tab:Hierarchical}
\end{table}


Despite the disappointing results of our preliminary experiments using acoustic models trained on isolated utterances, we continue to further explore this approach building on prior textual results. We were seeking of a way to leverage the meaningful contextual signals that could be present in adjacent turns. We shifted the method in a hierarchical training framework where first, acoustic models were trained on the target segments using isolated utterances without conversational context. Subsequently, we fine-tuned the model to adapt to the surrounding conversation segments, thereby learning higher-level emotional cues that are context-dependent. Simultaneously, we train a parallel model from the same baseline checkpoint to serve as a comparison, ensuring our fine-tuning process contributes positively to the emotion prediction task.

The obtained results, detailed in Table \ref{tab:Hierarchical} demonstrate the limited gains achieved through hierarchical fine-tuning with concatenated context. Critically, all context-enhanced models fail to improve over the baseline wav2vec2 model at 76.2\% UA. This implies significant shortcomings in the concatenation-based context integration paradigm.

Although small improvements are achieved using the previous context with MWCE+CCFTE, the global hierarchical learning methodology provides insignificant improvements to acoustic modeling. These results reveal shortcomings compared to text-based modeling approaches.

In particular, the minimal gains from concatenating context features (CCFTE) reveal this technique inadequately incorporates conversational patterns. The embedding masking (MWCE) is somewhat more beneficial, but the context integration remains insufficient.

We furthermore tried other experiments which not yielded better results, these experiments where based on MFCC cues of the surroundings segments.

\section{Analysis of Prediction accuracy based on the previous segment's emotion}



\begin{figure}[h]
    \centering
    \includegraphics[width=0.48\textwidth]{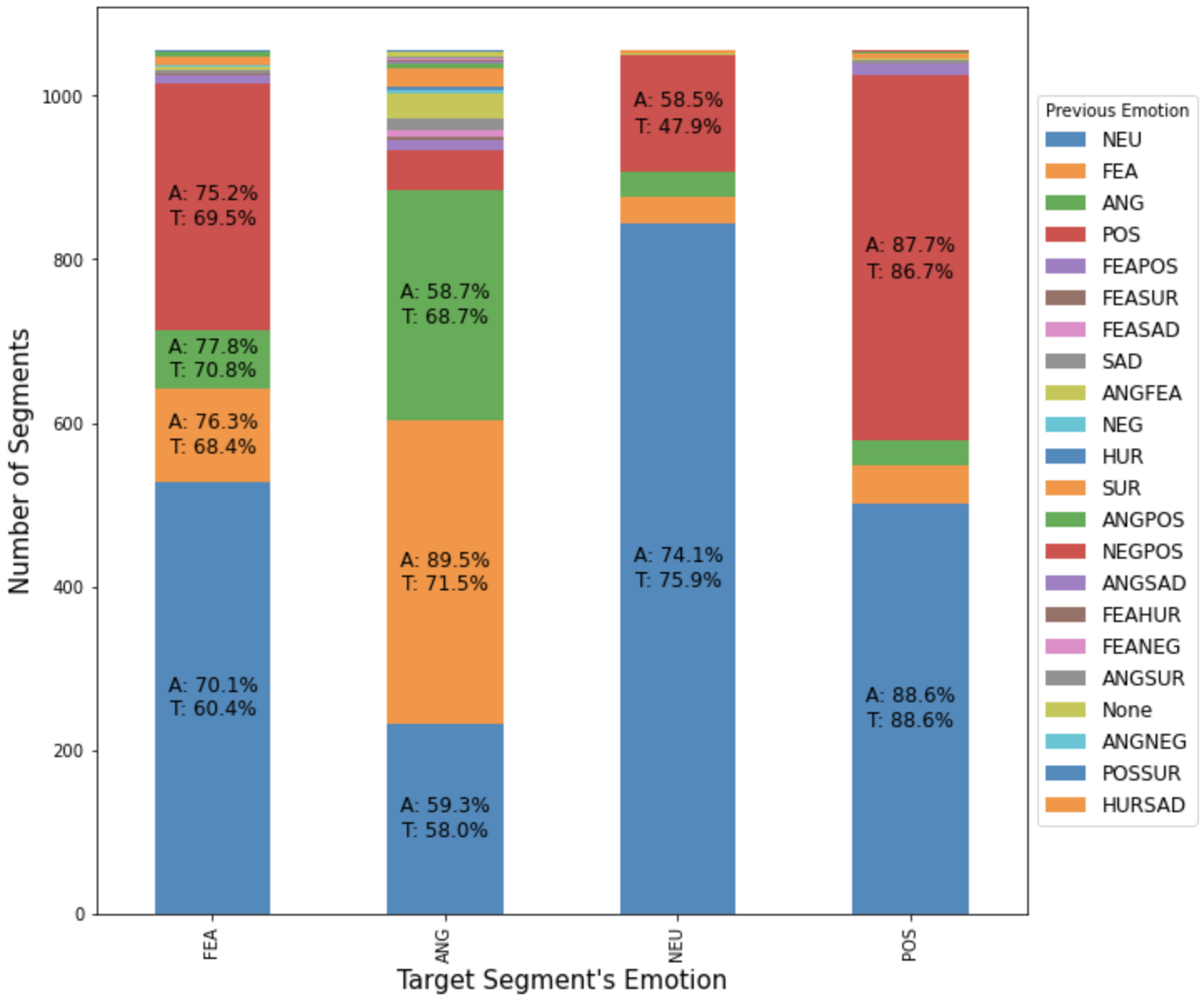}
    \caption{Prediction accuracy of the target emotions based on the previous segment's emotion, \% in UA}
    \label{fig:analysis}
\end{figure}

The Figure \ref{fig:analysis}, illustrates the distribution of previous emotion labels of the 4 targeted emotions. To compare the results obtained with conversational context, we took the same configuration with context taken from previous segments (whatever the speakers) for two different sets of predictions: speech transcriptions (T) and acoustic (A). Both models performance are respectively 71.2\% (T) and 76.2\% (A) UA (see Table \ref{tab:context} and \ref{tab:audio_results}).

Across both experiments, the Positive emotion and Neutral state segments seem to be predicted most accurately when the previous emotion is also Positive, results from 86.7\% to 88.6\% UA for both acoustic and transcriptions. The best results for Fear is obtained from Anger previous segment, 77.8\% (A) and 70.8\% (T). For Anger class an high UA is obtained for the segments with anterior Fear emotion expressed. The acoustic and textual models results are heterogeneous for the Anger class, the acoustic model is outperforming textual model when the previous segment was Fear (89.5\% (A) vs. 71.5\% (T)), on the other hand when the previous segment was Anger, the textual model had great results over the acoustic model (68.7\% (T) vs. 58.7\% (A)).

\section{Conclusion}

This paper explored Multiscale Contextual Learning for Speech Emotion Recognition in emergency call center conversations using the CEMO corpus collected in-the-wild. We conducted experiments incorporating contextual information from both speech transcriptions and acoustic signals with varying scales of the context. Overall, acoustic models demonstrate superior performance compared to text models, Table \ref{tab:context}, \ref{tab:Hierarchical}.

For text modeling with FlauBERT's Transformer embeddings, the context derived from previous segment has a more significant influence on accurate prediction than the following segment. Furthermore, taking the last speech turn of the same speaker in the conversation leads to better results in Table \ref{tab:context}. 

For acoustic modeling with wav2vec2.0 Transformer embeddings, we did not improve our results by using contextual information, Table \ref{tab:audio_results}. Despite pursuing a hierarchical training framework, Table \ref{tab:Hierarchical}, the results are disappointing and reveal challenges in effectively modeling sequential unimodal acoustic context using feature concatenation.

We also conducted an in-depth analysis of the impact of the previous emotions on the predictions. While multi-scale conversational context learning using Transformers can enhance performance in the textual modality for emergency call recordings, incorporating acoustic context is more challenging, see Table \ref{tab:audio_results}.
Advanced context modeling techniques are needed to fully leverage conversational dependencies in speech emotion recognition. Extending the context to model inter-speaker dynamics and relationships throughout full conversations is an important direction. Advances in attention mechanisms to handle wider contexts will also enable further progress on context-aware speech emotion recognition.

\subsection*{Ethics and reproducibility}
The use of the CEMO database or any subsets of it, carefully respected ethical conventions and agreements ensuring the anonymity of the callers. All evaluations are performed on 5 folds with a classical cross-speaker folding strategy that is speaker independent between training, validation and test sets. During each fold, system training is optimized on the best Unweighted Accuracy (UA) of the validation set. The outputs of each fold are combined for the final results. The experiments were carried out using Pytorch on GPUs (Tesla V100 with 32 Gbytes of RAM). To ensure the reproducibility of the runs, we set a random seed to 0 and prevent our system from using non-deterministic algorithms. This work was performed using HPC resources from GENCI–IDRIS (Grant 2022-AD011011844R1).


\begin{acks}
The PHD thesis of Theo Deschamps-Berger is supported by the ANR AI Chair HUMAAINE at LISN-CNRS, led by Laurence Devillers and reuniting researchers in computer science, linguists and behavioral economists from the Paris-Saclay University. The data annotation work was partially financed by several EC projects: FP6-CHIL and NoE HUMAINE. The authors would like to thank, M. Lesprit and J. Martel for their help with data annotation. The work is conducted in the framework of a convention between the APHP France and the LISN-CNRS.
\end{acks}

\clearpage
\balance
\bibliographystyle{ACM-Reference-Format}
\bibliography{ICMI_workshop_23}


\end{document}